# Learning the Generalizable Manipulation Skills on Soft-body Tasks via Guided Self-attention Behavior Cloning Policy

Xuetao Li, Fang Gao, Jun Yu, Shaodong Li, Feng Shuang

*Abstract*—Embodied AI represents a paradigm in AI research where artificial agents are situated within and interact with physical or virtual environments. Despite the recent progress in Embodied AI, it is still very challenging to learn the generalizable manipulation skills that can handle large deformation and topological changes on soft-body objects, such as clay, water, and soil. In this work, we proposed an effective policy, namely GP2E behavior cloning policy, which can guide the agent to learn the generalizable manipulation skills from soft-body tasks, including pouring, filling, hanging, excavating, pinching, and writing. Concretely, we build our policy from three insights:(1) Extracting intricate semantic features from point cloud data and seamlessly integrating them into the robot's end-effector frame; (2) Capturing long-distance interactions in long-horizon tasks through the incorporation of our guided self-attention module; (3) Mitigating overfitting concerns and facilitating model convergence to higher accuracy levels via the introduction of our two-stage fine-tuning strategy. Through extensive experiments, we demonstrate the effectiveness of our approach by achieving the 1st prize in the soft-body track of the ManiSkill2 Challenge at the CVPR 2023 4th Embodied AI workshop. Our findings highlight the potential of our method to improve the generalization abilities of Embodied AI models and pave the way for their practical applications in real-world scenarios. All codes and models of our solution is available at https://github.com/xtli12/GP2E.git.

*Note to Practitioners*—This paper explores the challenge of training artificial agents to execute complex manipulation tasks with soft-body objects, characterized by significant deformations and topological transformations. Traditional methods falter with materials such as clay, water, and soil, primarily due to the difficulty in generalizing manipulation skills across diverse scenarios. We present the GP2E behavior cloning policy, specifically developed to facilitate the learning of these skills in tasks including pouring, filling, hanging, excavating, pinching, and writing. The GP2E policy equips robots to capture essential long-distance interactions for managing complex, long-duration tasks efficiently, and concurrently mitigates overfitting and enhances model convergence, thus improving accuracy in task execution. Our findings indicate that the GP2E policy substantially improves the generalization capabilities of Embodied AI models, thereby broadening the prospects for their practical deployment in real-world settings.

Xuetao Li, Fang Gao, Shaodong Li, and Feng Shuang are with Guangxi Key Laboratory of Intelligent Control and Maintenance of Power Equipment, School of Electrical Engineering, Guangxi University, Nanning 530004, China (email: xtli312@163.com, fgao@gxu.edu.cn, lishaodongyx@126.com, fshuang@gxu.edu.cn) *(Corresponding author: Fang Gao.)*.

Jun Yu is with Department of Automation, University of Science and Technology of China, Hefei 230027, China (email: harryjun@ustc.edu.cn)

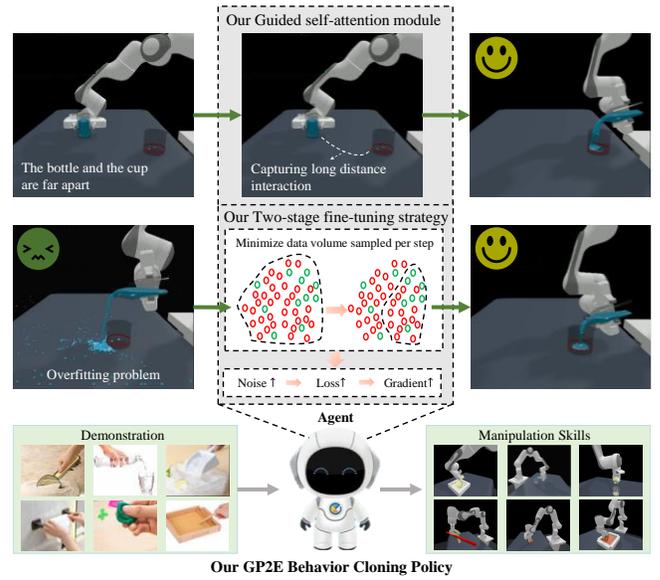

Fig. 1. Overview of our policy. By employing our Guided Point Cloud to End-effector (GP2E) behavior cloning policy, agents can learn the generalizable manipulation skills akin to those possessed by humans.

## I. INTRODUCTION

WITH the rise of Chat-GPT [1], AI (artificial intelligence) has once again sparked a global frenzy. But models like GPT, do not have a physical body to interact with the physical or virtual environments. In contrast, Embodied AI represents a significant advancement by integrating physical bodies into AI systems. These embodied agents gather environmental information through sensors and execute physical actions using mechanical actuators. Diverging from traditional AI approaches, which often rely on abstract symbolic manipulation or passive learning from static datasets, Embodied AI emphasizes the fusion of sensorimotor experiences with learning and decision-making processes. By imbuing intelligence in agents capable of perceiving, acting upon, and manipulating their surroundings, Embodied AI aims to develop systems with human-like understanding, reasoning, and behavior.

Embodied AI holds promise for automating various daily tasks, including household chores. To achieve this vision, robots must possess human-like manipulation skills, allowing them to manipulate diverse objects with ease after being trained on a variety of examples. Yet, many existing Embodied AI models rely heavily on extensive interactions with training environments, which may not be practical in real-world scenarios. To this end, the SAPIEN ManiSkill1 [2] introduced a comprehensive simulation benchmark for manipulating 3D objects. This benchmark leverages large-scale datasets of demonstrations to train agents and evaluates their generalization capabilities across various tasks, such as pushing chairs, opening cabinet doors, and moving buckets. Building upon this foundation, ManiSkill2 [3] further enhances the benchmark by incorporating a broader range of manipulation tasks to address the generalizability issue. However, despite these advancements, the baseline of ManiSkill2 still face limitations in performing Soft-body tasks, including pouring, filling, hanging, excavating, pinching, and writing.

Upon conducting a thorough investigation, we have identified several critical challenges that impede the baseline performance of ManiSkill2 in learning generalizable manipulation skills for Soft-body objects: 1) The baseline of ManiSkill2 relies on PointNet [4] to perceive the environment for action planning. However, this method is not robust enough to achieve generalizable shape understanding across complex topologies and geometries; 2) The long-horizon tasks featured in ManiSkill2 entail numerous long-distance interactions between objects and the robot. Successfully tackling these tasks requires the ability to effectively capture such interactions; 3) The demonstration trajectories provided for each task in ManiSkill2 are limited in quantity. Consequently, there exists a risk of overfitting during the training process, where the model may excessively adapt to the specific demonstration data rather than learning generalizable manipulation skills.

To address the challenges outlined above, inspired by [5], [6], [7], [8], [9], [10], [11], we present an effective policy termed the Guided Point cloud to End-effector (GP2E) behavior cloning policy (refer to Fig. 1), designed to facilitate the learning of generalizable manipulation skills from Soft-body tasks featured in the ManiSkill2 Challenge[1]. Our technical contributions encompass:

1) We propose an advanced 3D computer vision network architecture capable of extracting intricate semantic features from point cloud data and seamlessly integrating them into the robot's end-effector frame;
2) We propose a novel Guided self-attention module tailored to capture long-distance interactions between objects and the robot within long-horizon tasks;
3) We propose a Two-stage Fine-tuning Strategy aimed at mitigating overfitting concerns and facilitating model convergence to higher accuracy levels.

Our proposed method yields significant improvements in success rates, surpassing the ManiSkill2 baseline by an average of 18% across six Soft-body tasks. Notably, our method achieved first place in the "No Restriction (Soft Body)" track of the ManiSkill2 Challenge at the CVPR 2023 4th Embodied AI workshop[2].

[1] https://sapien.ucsd.edu/challenges/maniskill
[2] https://embodied-ai.org/#challenges

## II. RELATED WORK

### A. Soft-body Tasks

In real-world scenarios, robots encounter not only rigid bodies but also various types of soft materials such as cloth, water, and soil. Several simulators have been developed to facilitate robotic manipulation involving soft bodies. For instance, MuJoCo [12] and Bullet [13] utilize the finite element method (FEM) to simulate objects like ropes, cloth, and elastic materials. However, FEM-based approaches struggle with handling significant deformation and topological changes, such as scooping flour or cutting dough. Other environments, such as SoftGym [14] and ThreeDWorld [15] leverage Nvidia Flex to simulate large deformations, but they fall short in realistically simulating elasto-plastic materials like clay. PlasticineLab [16] employs the continuum-mechanics-based material point method (MPM), yet it lacks the capability to integrate with rigid robots and requires improvements in simulation and rendering performance. ManiSkill2 develops a custom GPU-based MPM simulator from scratch utilizing Nvidia's Warp [17] JIT framework and native CUDA for optimal efficiency and customization, ManiSkill2 is the first embodied AI environment to support 2-way coupled rigid-MPM simulation and the first to offer real-time simulation and rendering of MPM materials.

### B. Learning Generalizable Manipulation Skills

Generalizable manipulation skills are fundamental in the field of Embodied AI, empowering agents to tackle long-horizon and intricate daily tasks [18], [19]. Previous research endeavors have concentrated on discerning crucial components or extracting features of articulations to establish representations that facilitate generalized manipulation across diverse instances [20], [21], [22]. These approaches often rely on visual cues, such as key location identification, pose estimation, or pretrained attention models. Moreover, control-based methods employing model prediction and generative planning techniques have been investigated to achieve robust and adaptable control over both familiar and novel objects [23], [24]. Imitation learning offers a viable solution to equip robots with a variety of manipulation capabilities [25]. RoboCook [26] introduced Graph Neural Networks (GNNs) with imitation learning to model tool-object interactions, integrating tool classification with self-supervised policy learning to devise manipulation plans. The Diffusion policy [27] fully unlocked the potential of diffusion models for visuomotor policy learning on physical robots. To foster interdisciplinary collaboration and ensure the reproducibility of research on generalizable manipulation skills, it is essential to establish a versatile and publicly accessible benchmark. In this regard, ManiSkill2 has constructed a benchmark capable of accommodating object-level variations in both topological and geometric attributes, while also addressing the practical challenges inherent in manipulation tasks.

### C. Transformer-Based Vision Backbones

The advent of transformer-based models [28], [29], [30] in computer vision has marked a significant advancement since the introduction of the Transformer [31] architecture. These models surpassed convolutional networks in terms of both speed and accuracy in many down-stream tasks. However, conventional transformer-based networks often involve numerous dot-product operations between feature maps, leading to high computational demands. In our guided self-attention module, we strategically perform dot-product operations a single time, thereby achieving high accuracy while substantially reducing computational overhead.

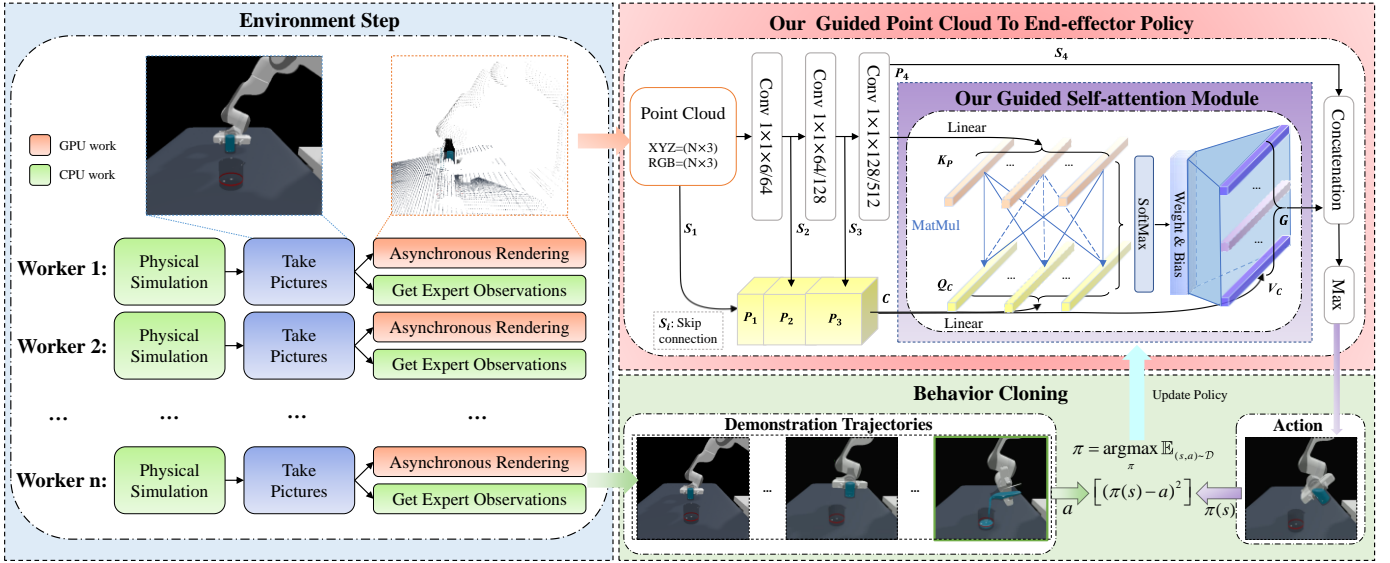

Fig. 2. The pipeline of our approach. Initially, environment samples are gathered from multiple workers. Subsequently, our GP2E (Guided Point Cloud To End-effector) policy seamlessly translates point cloud data into actions. Next, the Behavior Cloning algorithm guides our policy towards actions found within successful demonstrations. Finally, we employ our two-stage fine-tuning strategy to address overfitting concerns and aid model convergence towards higher accuracy levels.

*D. Point Cloud-Based Manipulation Policies*

Point cloud-based manipulation policies have been rigorously explored [32], [33], [34]. The FrameMiner [35] investigated how different coordinate frames for input point clouds affected manipulation skill learning in 3D environments, and proposed a dynamic frame selection method that adaptively merged the advantages of different frames, thereby enhancing performance in complex manipulation tasks without the need for modifying camera setups. Besides, researchers have also begun integrating point clouds into deep reinforcement learning (RL) frameworks to enhance manipulation learning [36], [37]. The process of feature learning within 3D neural networks introduces beneficial inductive biases for visual representation, leading to the development of a robust algorithm that outperforms traditional 2D approaches in intricate robotic manipulation tasks where precise encoding of relational dynamics is essential [38]. Consequently, our research prioritizes point cloud-based policies to facilitate the learning of generalized manipulation skills in robots.

## III. PROBLEM FORMULATION

Our problem focuses on policy learning for the development of generalizable soft-body manipulation skills in robots. This entails enabling robots to manipulate a diverse range of soft-body objects in conjunction with rigid bodies within a specific task domain. For instance, pouring water into cups positioned variably and with distinct final target liquid levels. The tasks and the real-time soft-body environments come from the ManiSkill2 Challenges. Consequently, our environment consists of a robot, a soft-body object coupled with rigid bodies (e.g., a water-filled rigid bottle), and multiple depth cameras. These cameras enable us to generate various single fused point clouds, which are then concatenated together to form the observations of the environment. The task goal is defined by the point cloud data. For example, in the task of pouring water, the target cup along with its final liquid levels are labeled within the water-filled cup. The task is deemed successful when the final liquid level precisely aligns with a designated target line. Assuming the robot state remains consistently known, our state, $S_t$, thus consists of a point cloud with labels assigned to individual points, alongside the robot state.

In addressing the challenge of cultivating generalizable soft-body manipulation skills in robots, we have introduced an effective policy framework. Our proposed approach, the guided self-attention based policy, adeptly captures highly condensed semantic features from point cloud data and seamlessly transforms these features into the end-effector frame of robots. The action generated by our policy on the state $s$ is denoted as $\pi(s)$, We employ a behavior cloning strategy to guide our policy $\pi$ towards favoring actions $a$ contained within successful demonstrations $\mathcal{D}$, achieved by minimizing the Euclidean distance between $\pi(s)$ and $a$:

$$\pi = \underset{\pi}{\operatorname{argmax}}\, \mathbb{E}_{(s,a)\sim\mathcal{D}}\left[(\pi(s)-a)^2\right] \quad (1)$$

## IV. METHOD

We now introduce our method. The objective of our methodology is to acquire generalizable soft-body manipulation skills capable of effectively addressing various soft-body tasks through a robust visual-to-end-effector policy. As depicted in Fig. 2, our methodology delineates its objective into three pivotal insights:

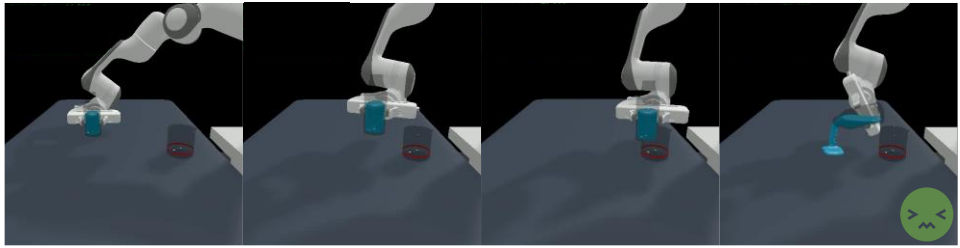
(a) Visualized video of **ManiSkill2** baseline on pouring task

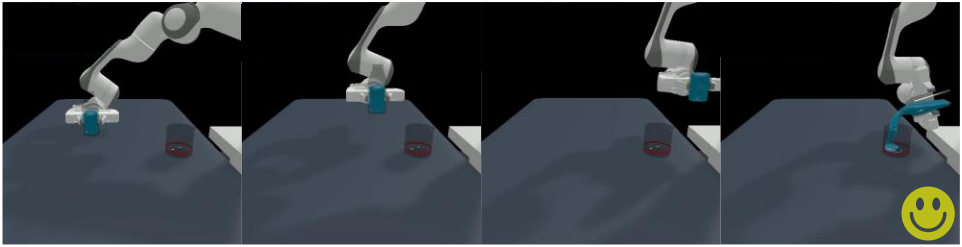
(b) Visualized video of **our method** on pouring task

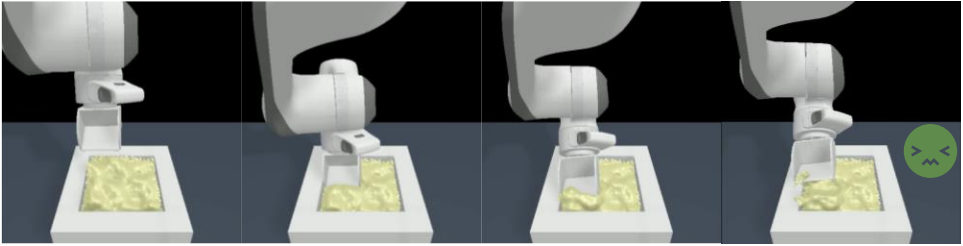
(c) Visualized video of **ManiSkill2** baseline on excavating task

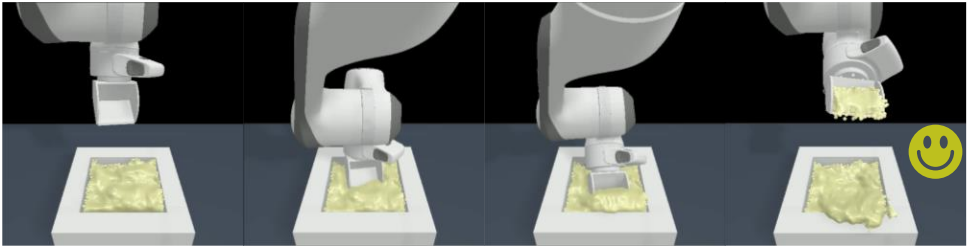
(d) Visualized video of **our method** on excavating task

Fig. 3. Visualized video of our method and ManiSkill2.

Extracting intricate semantic features from point cloud data and seamlessly integrating them into the robot's end-effector frame.
1) Capturing long-distance interactions in long-horizon tasks through the incorporation of our guided self-attention module.
2) Mitigating overfitting concerns and facilitating model convergence to higher accuracy levels via the introduction of our two-stage fine-tuning strategy.

*A. Guided Point Cloud To End-effector Policy*

In our pipeline, we adhere to the physical simulation and rendering procedures outlined by ManiSkill2: (1) Conducting physical simulation across multiple worker processes; (2) Taking pictures using both the base camera and the hand camera.; (3) Employing asynchronous rendering to convert images into point clouds on the GPU while simultaneously acquiring expert observations from the replay buffer on the CPU. Unlike ManiSkill1, where the CPU remains idle during GPU rendering, ManiSkill2 enhances CPU utilization by initiating expert observations while the GPU is engaged in rendering. This technique is named Asynchronous Rendering by ManiSkill2. Expert observations refer to trajectories that successfully accomplish tasks, serving as invaluable resources to facilitate learning-from-demonstrations methodologies.

Following the environment step in Maniskill2, we acquire two single fused point clouds from different cameras. Subsequently, we concatenate these points and remove ground artifacts using height clipping. In the baseline approach of ManiSkill2, PointNet [4] serves as the visual backbone to randomly downsample the point cloud to 1200 points. However, we observed that this baseline fails to capture intricate semantic features in tasks with long-horizon tasks.

To fully leverage the relative positional relationships between objects and the robot within the point cloud, we implement a method of reusing point cloud features from various levels by introducing skip connections and concatenating them into channel-wise condensed features. Additionally, we introduce guided self-attention to capture long-distance mapping relationships between objects and the robot. These features are then concatenated with the channel-wise up-sampled features and subjected to a softmax function to select suitable point cloud features for guiding the robot's movements.

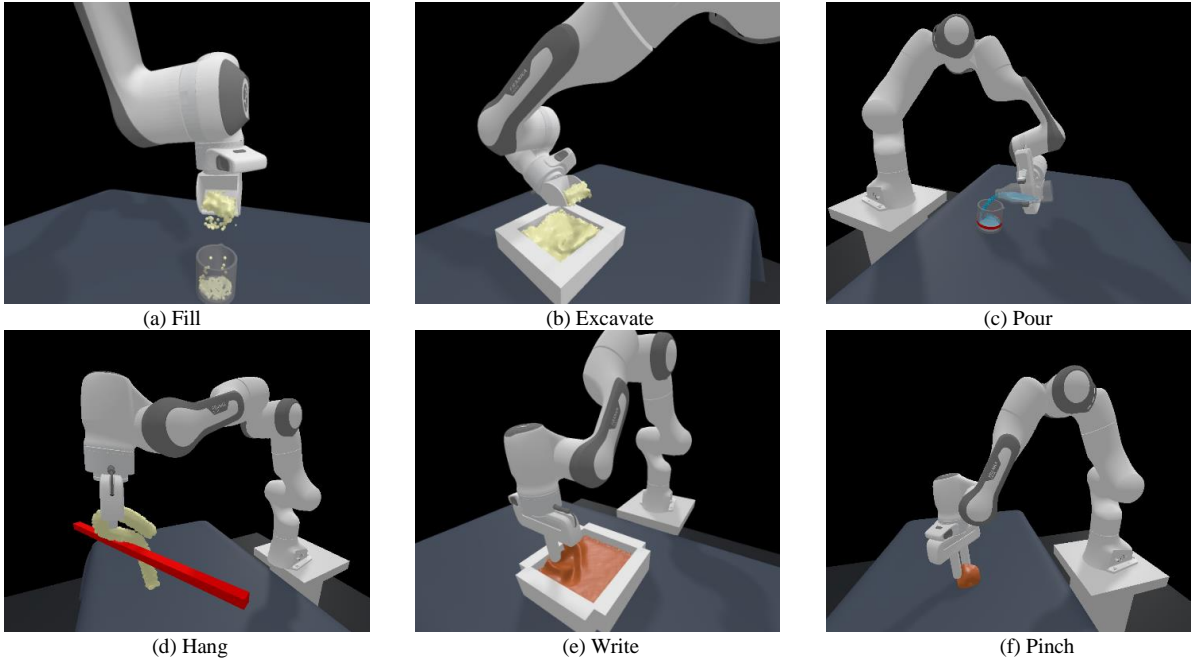

(a) Fill  (b) Excavate  (c) Pour
(d) Hang  (e) Write  (f) Pinch

Fig. 4. Tasks in Maniskill2 challenge.

As depicted in Fig. 2, the input of our point cloud to end-effector policy is a fused point cloud $P_1$ with 6 channels. Among these channels, 3 channels contain XYZ position information, while the remaining 3 channels encompass RGB information. To generate features with different mapping levels, we employ three Conv1×1 operations with varying input and output channel sizes: 6/64, 64/128, and 128/512, respectively. Following each Conv1×1 operation, we apply layer normalization and ReLU activation functions. The outputs of these operations are denoted as $P_2$, $P_3$, and $P_4$, respectively. Subsequently, we concatenate $P_1$, $P_2$, and $P_3$ to obtain $C$:

$$C = concat(P_1, \sigma(\mu_1(\phi_1(P_1))), \sigma(\mu_2(\phi_2(P_2)))) , \quad (2)$$

where $\sigma(\cdot)$ and $\mu_i(\cdot)$ (where $i=1,2$) denote ReLU activation functions and layer normalization, respectively. $\phi_i(\cdot)$ (where $i=1,2$) signifies three Conv1×1 with distinct input and output channel sizes. Subsequently, we incorporate our guided self-attention module to capture long-distance interactions between objects and the robot. The resulting feature, denoted as $G$, is concatenated with $P_4$ and processed through a maximum value selection function to derive the action $\pi(s)$ based on the current state $s$, thereby guiding the robot's movements:

$$\pi(s) = max(concat(G, P_4)) \quad (3)$$

After being processed by our point cloud to end-effector policy, our baseline can extract intricate semantic features and can seamlessly transform them into the robot's end-effector frame.

### B. Guided self-attention module

We have observed that in scenarios where the object is distant from the robot, the baseline model in ManiSkill2 fails to capture long-distance interactions between the object and the robot. This deficiency may lead to erroneous actions, as illustrated in Fig. 3(a), when the cup is positioned at the right edge of the table, the robot's front joint tuning direction restricts its ability to rotate the bottle clockwise. Consequently, the robot must relocate to the right side of the cup to pour water into it, necessitating the visual network backbone to effectively process long sequence information. However, the baseline model of ManiSkill2 appears to struggle in handling such situations. Additionally, in tasks such as Excavation, where a robot must excavate a specific volume of clay in a box, a simple yet effective visual network backbone is required for depth estimation. The robot must infer joint tuning angles based on sequential images of the state, necessitating the visual network backbone to capture long-distance interactions in the sequential images. However, the baseline model of ManiSkill2 appears inadequate in addressing this scenario effectively (refer to Fig. 3 (c)).

To capture long-distance interactions in long-horizon tasks, we propose the guided self-attention module. In the conventional self-attention mechanism [31], the query, keys, and values are all vectors linearly mapped by the same feature map, which is inefficient in local feature extraction [39]. In our guided self-attention module, we address this by introducing a highly condensed feature (referred to as feature $C$ in Fig. 2, containing 704 channels) as the query and value vectors (denoted as $Q_C$ and $V_C$ in Fig. 2). These vectors encompass rich local features that have been processed through convolutions at various mapping levels. Subsequently, we introduce $P_4$ (see Fig. 2) as the key vectors (see vectors $K_P$ in Fig. 2) to compute cosine similarities with every vector in $Q_C$. This process enables the learning of long-distance interactions between features that are distant in the feature map and allows for captur-ing changes in vectors across sequential images. Next, we apply a softmax function to these cosine similarities to derive weights for $V_C$. The formulation of the guided self-attention can be described as follows:

$$G = \text{softmax}\left(\frac{Q_C K_P^T}{\sqrt{d_k}} + B\right) V_C, \qquad (4)$$

where $d_k$ indicates the dimension of queries and keys, and $B$ is a relative position bias [28]. With the integration of our guided self-attention module, our method can effectively capture long-distance interactions, thereby enhancing its suitability for the long-horizon tasks (see Fig. 3(b), (d)).

*C. Behavior Cloning*

Due to the soft-body simulator in ManiSkill2 being tailored for visual learning environments, it is preferable to employ a straightforward yet efficient supervised learning algorithm. Specifically, matching the predicted action with the demonstrated action based on visual observations proves effective [40]. Among the spectrum of learning-from-demonstrations algorithms, behavior cloning stands out as a straightforward choice, requiring fewer resources to implement [2]. Hence, we adopt a behavior cloning strategy, aiming to directly match predicted and ground truth actions by minimizing the Euclidean distance.

*D. Two-stage Fine-tuning Strategy*

Lastly, we propose a two-stage fine-tuning strategy to assist the model in alleviating overfitting concerns and achieving higher levels of accuracy.

As the training process progresses, we have observed potential overfitting issues, wherein tasks that the model previously solved successfully may become unmanageable later on (refer to the First Stage in Fig. 2). This phenomenon may arise due to the model focusing excessively on certain scenarios within the dataset, thereby hindering its ability to generalize effectively to diverse scenarios. Additionally, the loss may become too small to produce substantial gradients necessary for converging to higher levels of accuracy.

To address these challenges, we propose a two-stage fine-tuning strategy aimed at introducing more variability into the training process to promote convergence to higher accuracy levels. Specifically, we reload the best-performing model from the first stage and then reduce the batch size and simulation steps per environment step during training to decrease the volume of data sampled per step. By reducing the data volume sampled per step, we introduce more noise into the training process, leading to larger losses and gradients. Consequently, the model can escape local minima and converge to higher levels of accuracy (refer to Fig. 2).

Through a series of experiments employing various scale strategies for batch size and simulation steps per environment step, we have discovered an intriguing result: utilizing a scale of 0.8 for batch size and 0.9 for simulation steps per environment step consistently leads to higher accuracy in tasks such as *Pour*, *Fill*, *Excavate*, and *Hang*. We firmly believe that our two-stage fine-tuning strategy holds promise for enabling researchers to delve deeper into other fields as well.

## V. EXPERIMENTS

*A. Datasets and evaluation metrics*

ManiSkill2 challenge includes 6 soft-body manipulation tasks that call for agents to engage with soft bodies (refer to Fig. 4), moving or deforming them to achieve predetermined target states.

*1) Fill*
- Objective: To transfer clay from a bucket into the target beaker.
- Success Metric: The task is successful when the volume of clay inside the target beaker exceeds 90% of its capacity, while maintaining the soft body velocity below 0.05.
- Evaluation Protocol: Conduct 100 episodes with varying initial rotations of the bucket and initial positions of the beaker.

*2) Hang*
- Objective: To hang a noodle on a target rod.
- Success Metric: Success is achieved when a portion of the noodle is positioned higher than the rod, both ends of the noodle rest on opposite sides of the rod, the noodle does not touch the ground, the gripper remains open, and the soft body velocity is maintained below 0.05.
- Evaluation Protocol: Conduct 100 episodes with varying initial positions of the gripper and rod poses.

*3) Excavate*
- Objective: To elevate a predetermined quantity of clay to a designated height.
- Success Metric: The task is considered successful when the lifted clay volume meets specified parameters, is positioned above a predefined height threshold, spillage is limited to fewer than 20 clay particles on the ground, and the soft body velocity is kept below 0.05.
- Evaluation Protocol: Conduct 100 episodes with varying bucket poses and initial clay heightmaps.

*4) Pour*
- Objective: To transfer liquid from a bottle into a beaker.
- Success Metric: Success is defined by ensuring that the liquid level in the beaker is within 4mm of the red line, spilled water is limited to fewer than 100 particles, the bottle returns to an upright position at the end of the task, and the robot arm velocity remains below 0.05.
- Evaluation Protocol: Conduct 100 episodes with varying bottle positions, water levels in the bottle, and beaker positions.

*5) Pinch*
- Objective: To mold plasticine into a predefined target shape.
- Success Metric: The task is successful when the Chamfer distance between the current plasticine shape and the target shape is less than 0.3 times the Chamfer distance between the initial shape and the target shape.
- Evaluation Protocol: Conduct 50 episodes with varying target shapes.

*6) Write*
- Objective: To inscribe a specified character onto clay. The target character is randomly selected from an alphabet containing over 50 characters.

TABLE I
MODEL PERFORMANCE AND ABLATION STUDIES ON THE SIX SOFT-BODY TASKS OF MANISKILL2 CHALLENGE. **#BC:** THE BEHAVIOR CLONING ALGORITHM,
#P: THE POINTNET, #F: OUR TWO-STAGE FINE-TUNING STRATEGY, #G: OUR GUIDED SELF-ATTENTION MODULE

| Method | #BC | #P | #F | #G | Fill↑ | Hang↑ | Excavate↑ | Pour↑ | Pinch↑ | Write↑ | Average↑ |
|---|---|---|---|---|---|---|---|---|---|---|---|
| I | ✓ | ✓ | | | 0.64±0.02 | 0.67±0.02 | 0.09±0.02 | 0.10±0.04 | 0.00±0.00 | 0.00±0.00 | 0.25±0.02 |
| II | ✓ | ✓ | ✓ | | 0.75±0.02 | 0.71±0.01 | 0.24±0.01 | 0.14±0.01 | 0.00±0.00 | 0.00±0.00 | 0.31±0.01 |
| III | ✓ | | | ✓ | 0.82±0.02 | 0.74±0.03 | 0.17±0.01 | 0.26±0.01 | 0.01±0.01 | 0.00±0.00 | 0.33±0.01 |
| IV | ✓ | | ✓ | ✓ | **0.95±0.02** | **0.87±0.01** | **0.39±0.02** | **0.33±0.03** | **0.01±0.01** | 0.00±0.00 | **0.43±0.02** |

TABLE II
MODEL PERFORMANCE OF OURS, MANISKILL2 BASELINE, AND THE SOTA
IMITATION LEARNING METHODS, AS WELL AS THE SECOND (CHENBAO) AND
THIRD PLACE (DEE) IN MANISKILL2 CHALLENGE ON THE SIX SOFT-BODY
TASKS OF THE CHALLENGE. AVG: AVERAGE. EXCA: EXCAVATE

| Method | Fill | Hang | Exca | Pour | Pinch | Write | **Avg** |
|---|---|---|---|---|---|---|---|
| Maniskill2 [3] | 0.45 | 0.35 | 0.08 | 0.02 | 0.00 | 0.00 | 0.15 |
| Dee | 0.14 | 0.38 | 0.14 | 0.00 | 0.00 | 0.00 | 0.11 |
| ChenBao | 0.50 | 0.28 | 0.00 | 0.06 | 0.00 | 0.00 | 0.14 |
| RoboCook [26] | 0.50 | 0.69 | 0.14 | 0.01 | 0.00 | 0.00 | 0.22 |
| Diffusion [27] | 0.94 | 0.86 | 0.15 | 0.05 | 0.00 | 0.00 | 0.33 |
| **Ours** | **0.95** | **0.87** | **0.39** | **0.33** | **0.01** | 0.00 | **0.43** |

- Success Metric: Success is achieved when the Intersection over Union (IoU) between the current pattern and the target character exceeds 0.8.
- Evaluation Protocol: Conduct 50 episodes with varying target characters.

There are 200 demonstration trajectories for each task (except Pinch with 1550 trajectories).

*B. Implementation details*

The proposed method is implemented based on the Maniskill2 frame [3]. For optimization, we utilize the Adam optimizer with an initial learning rate set to 0.0003 and a batch size of 256, in accordance with the approach outlined in ManiSkill2. As for the controller, we implement the *pd-joint-delta-pos* in all tasks, which has been integrated into ManiSkill2. Additionally, the initial number of simulation steps per environment step is configured to 500. Our demonstration translation process adheres to the guidelines established by the ManiSkill2 benchmark.

*C. Results and Analysis*

The findings are consolidated in Table I and Table II. Table I present the outcomes of our investigations across six distinct tasks conducted over 100 trials, each initialized with three distinct random seeds. Table II compares the performance of our method against the ManiSkill2 baseline [3], current state-of-the-art (SOTA) imitation learning methods [26], [27], and the second (ChenBao) and third place (Dee) finishers in the ManiSkill2 Challenge across the six soft-body tasks of the Challenge. As elucidated in Table I, our method incorporates advanced techniques such as Behavior Cloning from Demonstrations, a Two-stage Fine-tuning Strategy, and a Guided Self-attention Module. Collectively, these enhancements enable our proposed policy (Method IV) to demonstrate superior performance, achieving an average accuracy of 43% across the evaluated tasks. Table II highlights that our proposed policy achieved the highest score in the ManiSkill2 Challenge, surpassing the Diffusion Policy and RoboCook by average margins of 10% and 21%, respectively, across the six tasks. Particularly noteworthy is the achievement of a 0 to 1 breakthrough on the *Pinch* task. Subsequently, we will conduct an in-depth analysis to discern the individual contributions of each introduced technique.

*1) Effect of Two-stage Fine-tuning Strategy*: From the data presented in Table I, it is evident that [ManiSkill2 with our Two-stage Fine-tuning Strategy] (Method II), exhibits a noteworthy enhancement in success rate, showcasing a 6% improvement on average across the six tasks compared to the ManiSkill2 baseline (Method I). Furthermore, our policy [Behavior Cloning with our Guided Self-attention module and Two-stage Fine-tuning Strategy] (Method IV) demonstrates a significant boost in success rate, with a commendable 10% improvement on average across the six tasks when compared to [our policy lacking the Two-stage Fine-tuning Strategy] (Method III). In Fig. 5, the accuracy curves depicting the performance with the Two-stage Fine-tuning Strategy across various tasks are illustrated. Each evaluation point is derived from 100 episodes randomly selected with different random seeds. Notably, for the *Excavate* and *Pour* tasks, both Method I (depicted by the blue line) and Method III (depicted by the green line) exhibit a decline in accuracy following their top-1 accuracy points. This trend indicates the potential existence of overfitting issues in Method I and Method III. Through the application of our Two-stage Fine-tuning Strategy, both Method II and Method IV are able to introduce additional noise in certain gradient steps. Consequently, this results in a larger loss derived from the policy actions and the demonstration trajectory, thereby amplifying the gradient of the subsequent training step. As depicted by the orange line and red line in Fig. 5, this strategy facilitates the policy in escaping local minima points and achieving convergence to higher accuracy levels.

*2) Effect of Guided self-attention module*: From Table I, it is evident that [Behavior Cloning] combined with the [Guided self-attention module] (Method III) outperforms the ManiSkill2 baseline, represented by [Behavior Cloning] com-

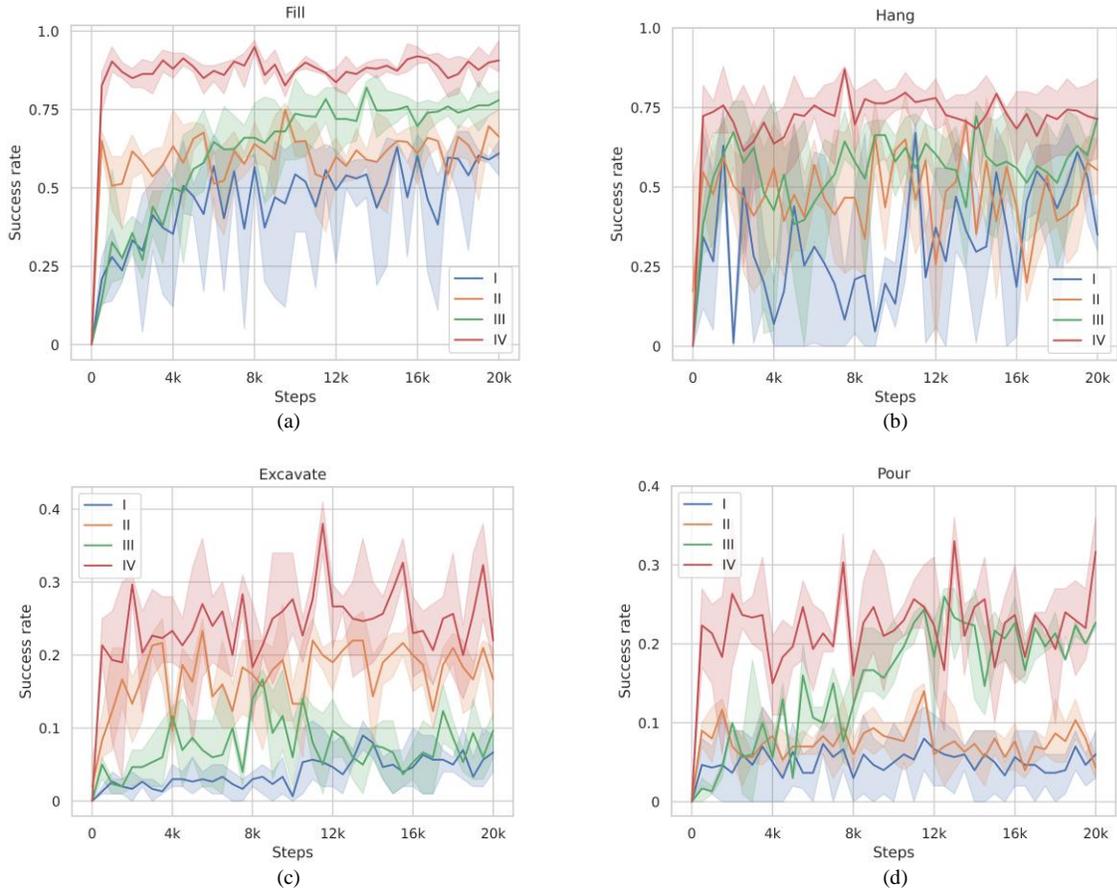

Fig. 5. The accuracy curve during training.

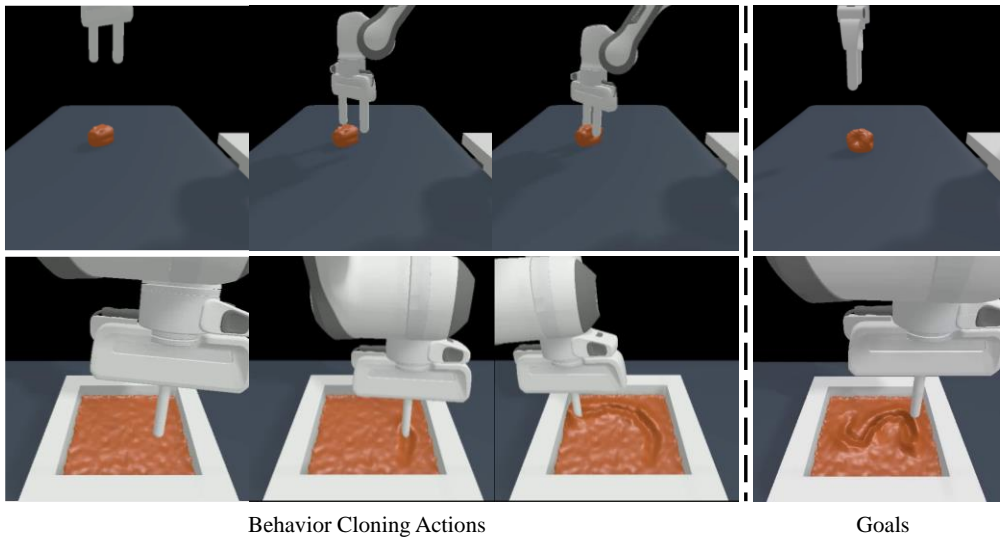

Behavior Cloning Actions          Goals

Fig. 6. Behavior cloning best actions for the tasks of *Pinch* and *Write*.

bined with [PointNet] (Method I), by 8% across the six tasks on average. The superiority of Method III can be attributed to our Guided self-attention module, which enables the model to capture more global information within the feature map. In essence, this means that the model equipped with our guided self-attention module can effectively extract long-distance interactions within the feature map. An illustrative example of the effectiveness of our guided self-attention module is depicted in Fig. 3 (a) and (b). When the target cup is positioned far away from the bottle, the ManiSkill2 baseline fails to capture the relationships between the cup and bottle, resulting in erroneous action sequences such as pouring water onto the table instead of into the target cup. In contrast, our guided self-attention module leverages skip connections to reuse previous feature maps in the network and computes cosine similarities between these reused feature maps and those that undergo straightforward channel-wise reshaping. Specifically, vectors containing the position information of the cup can be mapped to vectors containing the position information of the bottle by computing cosine similarities between them. Consequently, the agent can successfully match the bottle with the target cup. As depicted in Fig. 5, Method III (indicated by the green line) exhibits convergence to higher accuracy compared to Method I (indicated by the blue line), which does not incorporate our guided self-attention module.

*3) Further analysis of Soft-body tasks*: We note a distinct variance in the precision required across different tasks, which can lead to variations in accuracy scores even among tasks within the same category. For instance, tasks such as *Pour* and *Fill* both entail the manipulation of soft-body objects (liquid or clay) into a designated container. However, *Fill* exhibits a notably higher success rate compared to *Pour*. The underlying reason for this discrepancy lies in the precision demanded by each task. While *Fill* allows the robot agent to simply transfer all clay into the beaker, *Pour* necessitates a higher level of precision, specifically requiring the final liquid level to align precisely with a target line. Consequently, agents must meticulously control the tilt angle of the bottle to regulate the amount of liquid poured into the beaker accurately. Similarly, in the case of *Excavate*, agents must exercise keen judgment regarding the depth of excavation required to scoop up a specified quantity of clay. Conversely, tasks such as *Hang* do not mandate high-precision measurements from the agent, rendering them comparatively easier to accomplish. Furthermore, our observations indicate that Behavior Cloning agents struggle to effectively leverage target shapes to facilitate precise soft-body deformation. Notably, tasks such as *Pinch* and *Write*, which entail shape manipulation, present significant challenges for Behavior Cloning models, resulting in notably poor performance. As depicted in Fig. 6, while the robot learns the basic motion of pinching and demonstrates some progress toward the objective, the achieved level of proficiency falls short of the desired outcome. Similarly, in tasks such as Write, while the robot agent exhibits some capability in reproducing patterns, the resemblance to the target character remains insufficient.

## VI. CONCLUSION

In this paper, we address the challenges of overfitting in soft-body tasks by introducing our Two-stage Fine-tuning Strategy, and tackle the issue of capturing long-distance interactions through the implementation of our guided self-attention mechanism. We present a novel policy, the Guided Point Cloud to End-effector (GP2E) policy, which can seamlessly integrate the point cloud data into the robot's end-effector frame. Our experimental findings showcase that our methods yield notably higher success rates across six tasks when compared to existing baselines. Furthermore, our ablation studies validate the efficacy of each introduced technique.

REFERENCES

[1] T. Brown *et al.*, "Language models are few-shot learners," *Advances in Neural Information Processing Systems,* vol. 33, pp. 1877-1901, 2020.
[2] Mu, T., Ling, Z., Xiang, F., Yang, D. C., Li, X., Tao, S., ... & Su, H. "ManiSkill: Generalizable Manipulation Skill Benchmark with Large-Scale Demonstrations," *in Proc. Thirty-fifth Conference on Neural Information Processing Systems Datasets and Benchmarks Track (Round 2)*.
[3] Gu J, Xiang F, Li X, et al. ManiSkill2: A Unified Benchmark for Generalizable Manipulation Skills," *in Proc. Eleventh International Conference on Learning Representations*. 2022.
[4] C. R. Qi, H. Su, K. Mo, and L. J. Guibas, "Pointnet: Deep learning on point sets for 3d classification and segmentation," *in Proc. IEEE Conference on Computer Vision and Pattern Recognition*, 2017, pp. 652-660.
[5] Min Z, Zhu D, Ren H, *et al.*, "Feature-guided nonrigid 3-d point set registration framework for image-guided liver surgery: From isotropic positional noise to anisotropic positional noise," *IEEE Transactions on Automation Science and Engineering,* vol. 18, no. 2, pp. 471-483, 2020.
[6] Chen L, Xu Y, Zhu Q X, *et al.*, "Adaptive multi-head self-attention based supervised VAE for industrial soft sensing with missing data," *IEEE Transactions on Automation Science and Engineering*, 2023.
[7] He Y L, Li X Y, Xu Y, *et al.*, "Novel Distributed GRUs Based on Hybrid Self-Attention Mechanism for Dynamic Soft Sensing," *IEEE Transactions on Automation Science and Engineering*, 2023.
[8] S. Cheng and D. Xu, "League: Guided skill learning and abstraction for long-horizon manipulation," *IEEE Robotics and Automation Letters,* vol.8, no.10, pp: 6451-6458, 2023.
[9] J. Sun, L. Yu, P. Dong, B. Lu, and B. Zhou, "Adversarial inverse reinforcement learning with self-attention dynamics model," *IEEE Robotics and Automation Letters,* vol. 6, no. 2, pp. 1880-1886, 2021.
[10] T. Ablett, B. Chan, and J. Kelly, "Learning From Guided Play: Improving Exploration for Adversarial Imitation Learning With Simple Auxiliary Tasks," *IEEE Robotics and Automation Letters,* vol. 8, no. 3, pp. 1263-1270, 2023.
[11] H. Shen, W. Wan, and H. Wang, "Learning category-level generalizable object manipulation policy via generative adversarial self-imitation learning from demonstrations," *IEEE Robotics and Automation Letters,* vol. 7, no. 4, pp. 11166-11173, 2022.
[12] E. Todorov, T. Erez, and Y. Tassa, "Mujoco: A physics engine for model-based control," in *Proc. IEEE/RSJ International Conference on Intelligent Robots and Systems*, 2012: IEEE, pp. 5026-5033.
[13] Coumans E, Bai Y. Pybullet, "A python module for physics simulation for games," Robotics and Machine Learning, 2016). URL http://pybullet.org, 2016.
[14] X. Lin, Y. Wang, J. Olkin, and D. Held, "Softgym: Benchmarking deep reinforcement learning for deformable object manipulation," in *Conference on Robot Learning*, 2021, PMLR, pp. 432-448.
[15] Gan, Chuang, et al. "ThreeDWorld: A Platform for Interactive Multi-Modal Physical Simulation," *in Annual Conference on Neural Information Processing Systems*, 2021.
[16] Z. Huang *et al.*, "Plasticinelab: A soft-body manipulation benchmark with differentiable physics," *in Proc. International Conference on Learning Representations*, 2020.
[17] M. Macklin, "Warp: A high-performance python framework for gpu simulation and graphics," in *NVIDIA GPU Technology Conference (GTC)*, 2022.
[18] Wu H, Yan W, Xu Z, *et al.*, "A framework of robot skill learning from complex and long-horizon tasks," *IEEE Robotics and Automation Letters,* vol. 19, no. 4, pp. 3628-3638, 2021.
[19] J. Gu, D. S. Chaplot, H. Su, and J. Malik, "Multi-skill mobile manipulation for object rearrangement," *in Proc. The Eleventh International Conference on Learning Representations* , 2022.
[20] M. Mittal, D. Hoeller, F. Farshidian, M. Hutter, and A. Garg, "Articulated object interaction in unknown scenes with whole-body mobile manipulation," in *2022 IEEE/RSJ International Conference on Intelligent Robots and Systems (IROS)*, 2022: IEEE, pp. 1647-1654.
[21] M. Arduengo, C. Torras, and L. Sentis, "Robust and adaptive door operation with a mobile robot," *Intelligent Service Robotics,* vol. 14, no. 3, pp. 409-425, 2021.
[22] C. Devin, P. Abbeel, T. Darrell, and S. Levine, "Deep object-centric representations for generalizable robot learning," *in Proc. IEEE International Conference on Robotics and Automation (ICRA)*, 2018:

IEEE, pp. 7111-7118.
[23] B. Abbatematteo, S. Tellex, and G. Konidaris, "Learning to generalize kinematic models to novel objects," in *Proc. The 3rd Conference on Robot Learning*, 2019.
[24] A. Jain and S. Niekum, "Learning hybrid object kinematics for efficient hierarchical planning under uncertainty," in *2020 IEEE/RSJ International Conference on Intelligent Robots and Systems (IROS)*, 2020: IEEE, pp. 5253-5260.
[25] Mandlekar A, Xu D, Wong J, *et al.*, "What matters in learning from offline human demonstrations for robot manipulation," *arXiv preprint arXiv:2108.03298*, 2021.
[26] H. Shi, H. Xu, S. Clarke, Y. Li, and J. J. a. p. a. Wu, "Robocook: Long-horizon elasto-plastic object manipulation with diverse tools," in *Conference on Robot Learning (CoRL)*, 2023, pp: 642-660.
[27] Chi C, Feng S, Du Y, et al., "Diffusion policy: Visuomotor policy learning via action diffusion," in *Robotics: Science and Systems (RSS)*, 2023.
[28] A. Dosovitskiy *et al.*, "An image is worth 16x16 words: Transformers for image recognition at scale," *arXiv preprint arXiv:2010.11929,* 2020.
[29] H. Touvron, M. Cord, M. Douze, F. Massa, A. Sablayrolles, and H. Jégou, "Training data-efficient image transformers & distillation through attention," in *International Conference on Machine Learning*, 2021:PMLR, pp. 10347-10357.
[30] Z. Liu *et al.*, "Swin transformer: Hierarchical vision transformer using shifted windows," in *Proc. of the IEEE/CVF International Conference on Computer Vision*, 2021:IEEE, pp. 10012-10022.
[31] A. Vaswani *et al.*, "Attention is all you need," *Advances in Neural Information Processing Systems,* vol. 30, 2017.
[32] S James S, Davison A J, "Q-attention: Enabling efficient learning for vision-based robotic manipulation," *IEEE Robotics and Automation Letters*, vol. 7, no. 2, pp. 1612-1619, 2022.
[33] M. Sundermeyer, A. Mousavian, R. Triebel, and D. Fox, "Contact-graspnet: Efficient 6-dof grasp generation in cluttered scenes," in *2021 IEEE International Conference on Robotics and Automation (ICRA)*, 2021: IEEE, pp. 13438-13444.
[34] J. Lv, Q. Yu, L. Shao, W. Liu, W. Xu, and C. Lu, "Sagci-system: Towards sample-efficient, generalizable, compositional, and incremental robot learning," in *2022 International Conference on Robotics and Automation (ICRA)*, 2022: IEEE, pp. 98-105.
[35] Liu M, Li X, Ling Z, et al., "Frame mining: a free lunch for learning robotic manipulation from 3d point clouds," *arXiv preprint arXiv:2210.07442*, 2022.
[36] T. Chen, J. Xu, and P. Agrawal, "A system for general in-hand object re-orientation," in *Conference on Robot Learning*, 2022, PMLR, pp. 297-307.
[37] Huang W, Mordatch I, Abbeel P, et al., "Generalization in dexterous manipulation via geometry-aware multi-task learning," *arXiv preprint arXiv:2111.03062*, 2021.
[38] Ling Z, Yao Y, Li X, et al., "On the efficacy of 3d point cloud reinforcement learning," *arXiv preprint arXiv:2306.06799*, 2023.
[39] H. Lin, X. Cheng, X. Wu, and D. Shen, "Cat: Cross attention in vision transformer," in *2022 IEEE International Conference on Multimedia and Expo (ICME)*, 2022: IEEE , pp. 1-6.
[40] Liu X, Huang P, Liu Z, "A novel contact state estimation method for robot manipulation skill learning via environment dynamics and constraints modeling," *IEEE Transactions on Automation Science and Engineering*, vol. 19, no. 4, pp. 3903-3913, 2022.

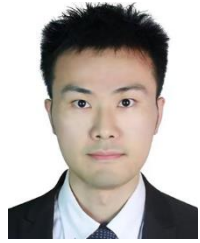

**Xuetao Li** received the B.S. degree from the Civil Aviation University of China, Tianjin, China. He is currently a Graduate Student at the School of Electrical Engineering, Guangxi University, Nanning, China. His research interests include computer vision and embodied artificial intelligence.

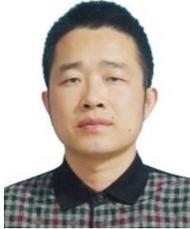

**Fang Gao** received the B.S. and Ph.D. degrees in Chemical Physics from University of Science and Technology of China in 2004 and 2010. After graduation, he worked as an Assistant Professor and an Associate Professor in the Institute of Intelligent Machines, Chinese Academy of Sciences. He is currently a Professor in the College of Electrical Engineering, Guangxi University, Nanning, Guangxi, China. His major research interests include multimedia computing, embodied artificial intelligence and quantum machine learning. His research work has been published in TMM, TCSVT, TSTE, RAL, ACM MM, etc. He has won champions from many Grand Challenges, such as CVPR 2023 Embodied AI Workshop ManiSkill2 Challenge and CVPR 2024 Embodied AI Workshop ManiSkill-ViTac Challenge, and the 2nd CCF Origin Pilot Cup Quantum Computing Challenge in the professional group.

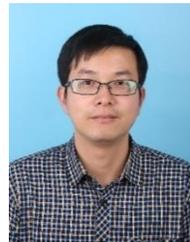

**Jun Yu** Jun Yu is currently an associate professor and laboratory director with the Department of Automation and the Institute of Advanced Technology, University of Science and Technology of China. His research interests are Multimedia Computing and Intelligent Robot. He has published 200+ journal articles and conference papers in TPAMI, IJCV, JMLR, TIP, TMM, TASLP, TCYB, TITS, TCSVT, TOMM, TCDS, ACL, CVPR, ICCV, NeurIPS, ICML, ICLR, MM, SIGGRAPH, VR, AAAI, IJCAI, etc. He has received 6 Best Paper Awards from premier conferences, including CVPR PBVS, ICCV MFR, ICME, FG, and won 50+ champions from Grand Challenges held in NeurIPS, CVPR, ICCV, MM, ECCV, IJCAI, AAAI.

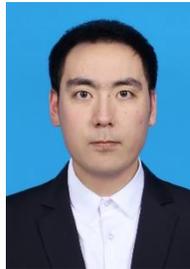

**Shaodong Li** received the M.S. degree in mechanical and electronic engineering from Northeastern University, Shenyang, China, in 2015. He received the Ph.D. degree in mechanical and electronic engineering from Harbin Institute of Technology, Harbin, China, in 2021. He is currently an assistant professor in Guangxi University, Nanning, China. His research interests include human-robot cooperation and robotic intelligent manipulation. His research work has been published in IEEE Robotics and Automation Letters, IEEE Transactions on Human-Machine Systems, Pattern Recognition Letters, etc.

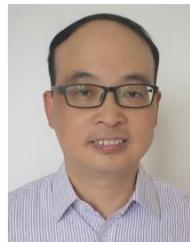

**Feng Shuang** received the B.S. degree from the Special Class of Gifted Young, University of Science and Technology of China (USTC), in 1995, and the Ph.D. degree from the Department of Chemical Physics, USTC, in 2000. He was a Research Associate with Princeton University, from 2001 to 2003, and was a Research Staff Member at Princeton University, from 2004 to 2009. In 2009, he joined the Institute of Intelligent Machines (IIM), China, as a Full Professor, and then was selected as a member of the One Hundred Talented People of Chinese Academy of Sciences. He is currently a Professor with Guangxi University. His research interests include system control and information acquisition, including multidimensional force sensors, quantum system control.